\documentclass[journal]{IEEEtran}

\ifCLASSINFOpdf
\usepackage[pdftex]{graphicx}
\usepackage{amsmath}
\usepackage{gensymb}
\usepackage{url}
\usepackage{hyperref}
\newcommand{\dvtm}{\(\Delta\)$v_{\mathrm{TM}}$ }
\newcommand{\stdv}{average steady velocity}

\usepackage{authblk}
\begin{document}
\title{SBTenv: A Literature-Validated Split-Belt Treadmill Simulation Environment}
\author[1]{Philip Byrne}
\author[1,*]{John F. Drazan}
\author[2,*]{Nidhi Seethapathi}

\affil[1]{Department of Electrical and Biomedical Engineering, Fairfield University}
\affil[2]{Department of Brain and Cognitive Sciences, Massachusetts Institute of Technology}
\affil[*]{These authors contributed equally}

\maketitle

\begin{abstract}
\boldmath
The split-belt treadmill is an important tool for studying human locomotor adaptation. Physics-based simulations complement such experiments by allowing hypotheses about the mechanisms of adaptation to be tested directly, and high-throughput simulation is increasingly needed as these models grow in complexity. However, no open-source split-belt treadmill environment exists in MuJoCo, a physics engine widely used for high-throughput simulation. Here, we present SBTenv, a split-belt treadmill environment built in MuJoCo and validated against published experimental data from a rimless wheel walking passively on a split-belt treadmill. By modeling compliant contacts at the spoke tips, SBTenv reproduces the experimental average steady velocity ($R^2=0.643$, RMSE $=0.0162$ m/s) without the empirical geometric correction required in prior simulations. We expect SBTenv, openly shared, to enable high-throughput simulation of musculoskeletal and robotic locomotion on a split-belt treadmill.
\end{abstract}


\IEEEpeerreviewmaketitle

\section{Introduction}
\IEEEPARstart{T}{he} split-belt treadmill (SBT) is an important tool for the study of human movement. In motor control and motor learning, the SBT is used to understand how locomotor symmetry adapts \cite{reisman_neurophysiologic_2010}\cite{roemmich_closing_2018}\cite{price_minimum_2023}. Computational models complement these experiments by allowing hypotheses about the mechanisms underlying split-belt adaptation to be tested directly, for example by comparing candidate control objectives or learning rules. While split-belt simulations have recently been developed \cite{price_minimum_2023}\cite{seethapathi2024exploration}\cite{fleischmann2025model}, we lack open-source environments for SBT simulation in high-throughput physics engines like MuJoCo \cite{todorov2012mujoco}. We therefore developed SBTenv, an \textit{in silico} SBT environment in MuJoCo that supports high-throughput simulation, provides a Gymnasium interface, and is validated against a published experimental dataset of a rimless wheel walking on a split-belt treadmill \cite{butterfield_split-belt_2022}.

Physics-based simulation has long been used to study locomotion, from passive dynamic walkers \cite{mcgeer_passive_1990}\cite{coleman_motions_1997} to optimization-based predictions of human gait \cite{anderson_dynamic_2001}\cite{srinivasan_computer_2006} and reflex-based neuromuscular control \cite{geyer_muscle-reflex_2010}. Deep reinforcement learning has since extended these approaches to high-dimensional musculoskeletal models \cite{song_deep_2021}\cite{schumacher_emergence_2025}, with applications ranging from identifying principles that reproduce natural gait \cite{schumacher_emergence_2025} to predicting the effects of clinical impairments \cite{ong_predicting_2019} and age-related falls \cite{nowakowski_deep_2022}, and to training controllers that transfer to legged robots \cite{hwangbo_learning_2019}. Because these methods require many simulated trials, they expose a tension between model fidelity and computational cost. MuJoCo \cite{todorov2012mujoco} enables such high-throughput simulation more than two orders of magnitude faster than OpenSim \cite{delp_opensim_2007}\cite{seth_opensim_2018} at comparable complexity, and existing tools convert OpenSim models into MuJoCo with close agreement in muscle kinematics and dynamics \cite{ikkala_converting_2021}\cite{wang_myosim_2022}. A split-belt environment in MuJoCo would therefore enable making these models available for studying split-belt locomotion and adaptation.
\begin{figure}[!t]
\centering
\includegraphics[width=\linewidth]{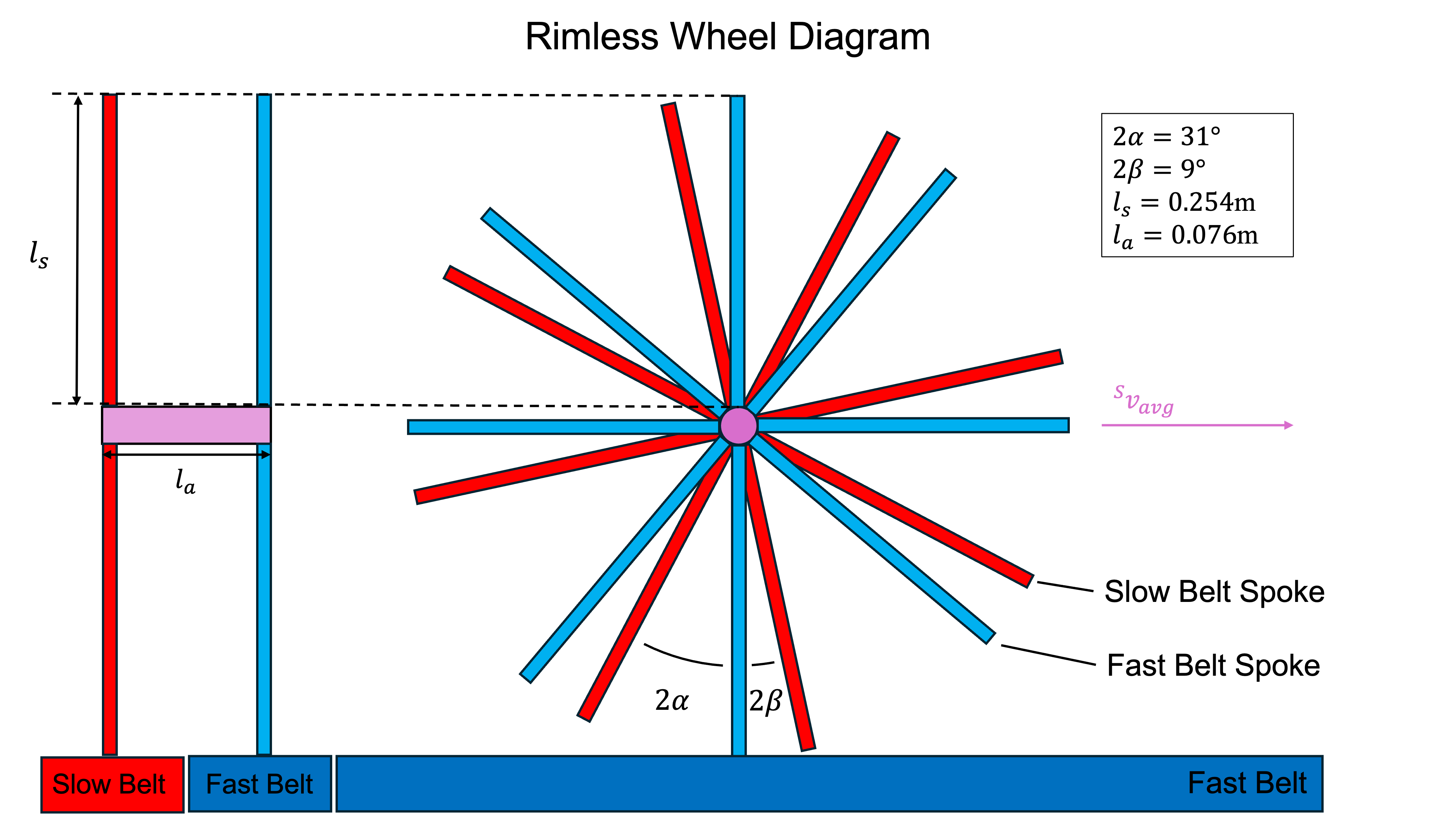}
\caption{Split-belt rimless wheel model in MuJoCo.}
\label{fig_sim}
\end{figure}

\section{Methods}
\subsection{Codebase}
SBTenv is written in the dm\_control Python framework (version $1.0.22$) \cite{tassa_dm_control_2020}, which uses the MuJoCo physics engine (version $3.2.2$) for physics simulation and reinforcement learning environments. We chose dm\_control for its direct access to MuJoCo and its ease of implementation. SBTenv provides a Gymnasium interface \cite{towers_gymnasium_2025} and is freely available on \href{https://github.com/PhilipByrn3/sbtenv}{GitHub}.

\subsection{Validation Approach}
We validated SBTenv using the split-belt rimless wheel model (SBWM) \cite{butterfield_split-belt_2022}, a rimless wheel (RW) that walks passively on an SBT by harvesting energy from the belt-speed difference (BSD). The SBWM is a minimal model of passive split-belt walking with published experimental data from a physical implementation. Because it is purely mechanical, with no musculoskeletal dynamics or controller, discrepancies between simulation and experiment can be attributed to the treadmill and contact model rather than to assumptions about biomechanics or control. Its simple, reproducible construction also allows other laboratories to build the same device on their own treadmills and use it to calibrate SBTenv, particularly its contact mechanics, to their setup. We refer to our simulated implementation as SBWM-env (with wheel RW-env) and to the physical implementation as SBWM-exp (with wheel RW-exp). SBWM-exp data reported graphically in \cite{butterfield_split-belt_2022} were digitized from figures using Fiji (version $1.54$p) \cite{schindelin_fiji_2012}.
\subsection{Split-Belt Treadmill Model}
SBTenv contains a slow belt and a fast belt, each modeled as a box geometry with a slide joint along the $y$-axis. The belts are separated by $2$ mm along the $x$-axis so that they do not contact each other. Neither slide joint is motor-driven; instead, belt motion is imposed kinematically. At the beginning of every timestep, each belt is assigned a constant velocity, while a narrow joint range keeps it from translating in space. This approximates a physical treadmill, in which the belt moves over two motor-driven rollers without the belt assembly itself translating. Each belt had a sliding friction coefficient of $1.15$, torsional and rolling friction coefficients of $0.1$, and a mass of $3.0$ kg.

\subsection{Rimless Wheel Model}
RW-env followed the specifications of RW-exp \cite{butterfield_split-belt_2022}: $9$ spokes per side, spoke length $0.254$ m, axle length $7.6$ cm, angular spacing between spokes of $40^\circ$, angular offsets $\alpha=15.5^\circ$ and $\beta=4.5^\circ$ ($2\alpha=31^\circ$, $2\beta=9^\circ$), and total mass $3.53$ kg (excluding rubber contacts). To simplify the experimental apparatus, RW-env was constrained to translate in the $y$--$z$ plane and rotate about its axle. Each simulation initialized RW-env at a height of $0.38$ m and, on the first timestep, set its velocity equal to the fast-belt velocity so that it could reach a steady state.

\subsection{Contact Model}
To capture the deformation and shock absorption of the rubber strips on RW-exp, we added a compliant contact at the tip of each RW-env spoke. Each contact was modeled as a sphere of radius $0.017$ m mounted on a slide joint at the spoke tip, with the joint acting as a spring-mass damper (range $0$--$0.015$ m, stiffness $50$ N/m, damping $10$ N$\cdot$s/m). The contacts had sliding and torsional friction coefficients of $1.0$ and a rolling friction coefficient of $0.1$. Because friction, stiffness, and damping values were not reported in \cite{butterfield_split-belt_2022}, we set them manually to suppress spoke bouncing on belt contact while allowing stable passive walking across the range of BSDs.

\subsection{Simulation Protocol}
SBWM-env was simulated at each digitized SBWM-exp BSD, with the slow belt stationary and the fast-belt velocity equal to the BSD. A simulation terminated when RW-env completed $12$ full rotations or after $2000$ timesteps. Average steady velocity, the validation metric, was computed as the axle displacement divided by elapsed time, measured from the end of a $0.05$ s settling period to the final timestep. All simulations used RK4 integration, the PGS constraint solver (maximum $50$ iterations), an elliptic friction cone, and a timestep of $0.01$ s.

\subsection{Rubber Contact and Offset Ablation}
To test whether the compliant contacts and the empirical fast-belt $z$-offset correction from \cite{butterfield_split-belt_2022} compensate for the same collision energy dissipation mechanism, we performed a full factorial ablation over contact placement and offset. Contact placement took four levels: none, fast-belt spokes only, slow-belt spokes only, and both (the configuration used in SBWM-env). Offset took two levels: none, and the $4.9$ mm fast-belt $z$-offset reported in \cite{butterfield_split-belt_2022}. All other parameters were held constant across the eight configurations, each of which was evaluated against the digitized SBWM-exp data using the metrics below.

\subsection{Statistical Analysis}
Agreement between SBWM-env and SBWM-exp was assessed by comparing average steady velocities point by point at each digitized BSD. Trials in which RW-env failed to complete one rotation were excluded. Agreement was quantified by
\begin{equation}
R^2 = 1-\frac{\sum_{i}(y_i-\hat{y}_i)^2}{\sum_{i}(y_i-\bar{y})^2}
\label{r2eq}
\end{equation}
where $y_i$ is the digitized SBWM-exp average steady velocity, $\hat{y}_i$ is the SBWM-env average steady velocity at the same BSD, and $\bar{y}$ is the mean of the SBWM-exp values. Because $R^2$ here measures agreement between two independently generated curves rather than the fit of a regression, it is unbounded below; negative values indicate a worse fit than the experimental mean. We also report the root-mean-square error (RMSE) and mean absolute error (MAE) between $y_i$ and $\hat{y}_i$. All analyses were performed in Python $3.10$ using NumPy \cite{harris_array_2020}.

\section{Results}
\subsection{Validation Against Experiment}
\begin{figure}[!t]
\centering
\includegraphics[width=\linewidth]{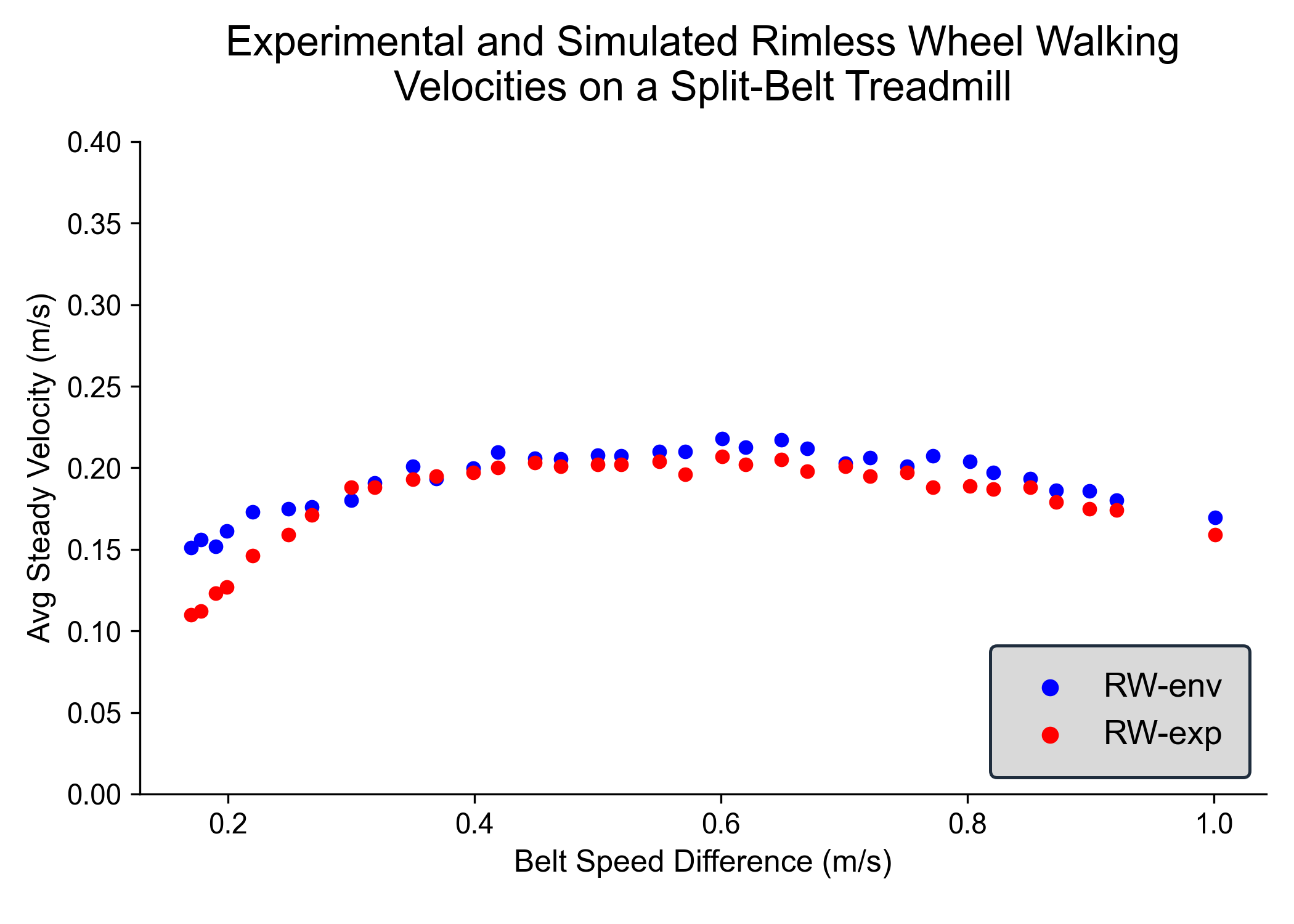}
\caption{Average steady velocity of the rimless wheel as a function of belt-speed difference in SBWM-env (both rubber contacts, no offset) and SBWM-exp \cite{butterfield_split-belt_2022}.}
\label{fig_validation}
\end{figure}

SBWM-env reproduced the dependence of RW average steady velocity on BSD observed in SBWM-exp (Figure \ref{fig_validation}), with $R^2=0.643$, RMSE $=0.0162$ m/s, and MAE $=0.0122$ m/s. Simulation required on average $0.213$ ms per Gymnasium \verb|env.step()| call (across $68{,}000$ physics timesteps) on a single core of an Apple M3, or roughly $4{,}700$ steps per second.

\subsection{Ablation}
\begin{figure*}[!t]
\centering
\includegraphics[width=0.75\linewidth]{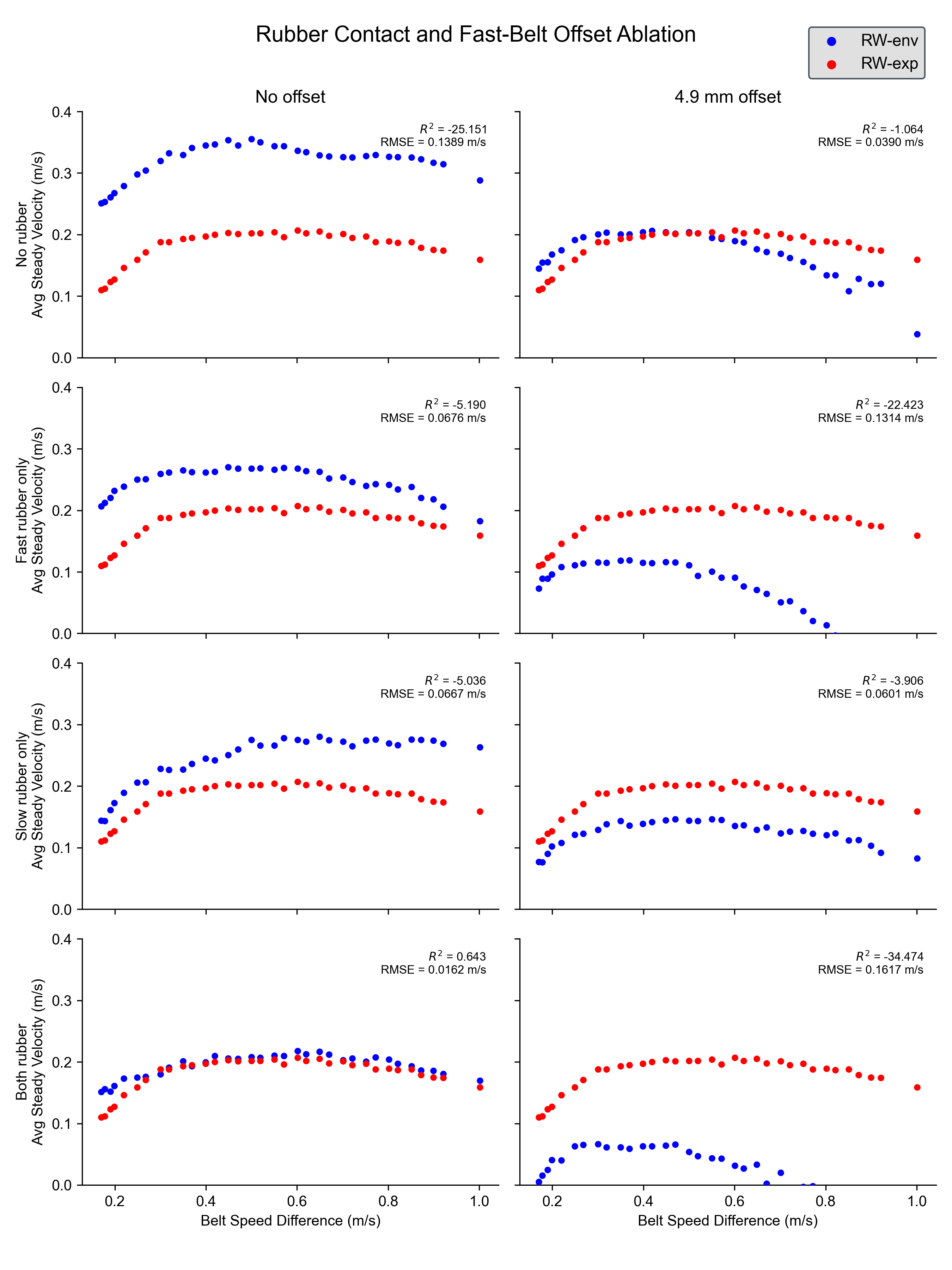}
\caption{Rubber contact and fast-belt offset ablation. Rows vary rubber contact placement and columns vary fast-belt offset. Blue points are SBWM-env average steady velocities from a single trial at each digitized BSD; red points are the digitized SBWM-exp values.}
\label{fig_ablation}
\end{figure*}

Table \ref{tab_ablation} and Figure \ref{fig_ablation} report agreement for all eight contact and offset configurations. The configuration used in SBWM-env (both contacts, no offset) gave the best fit ($R^2=0.643$), and the same contacts combined with the offset gave the worst ($R^2=-34.474$). Without the offset, fit improved as contacts were added. With the offset, fit worsened as contacts were added, and every configuration with contacts under-predicted the SBWM-exp average steady velocity.

\begin{table}[!t]
\centering
\caption{Rubber contact and offset ablation results.}
\label{tab_ablation}
\begin{tabular}{lccc}
\hline
Configuration & $R^2$ & RMSE (m/s) & MAE (m/s) \\
\hline
No rubber, no offset          & $-25.151$ & $0.1389$ & $0.1387$ \\
Fast rubber only, no offset   & $-5.190$  & $0.0676$ & $0.0648$ \\
Slow rubber only, no offset   & $-5.036$  & $0.0667$ & $0.0632$ \\
\textbf{Both rubber, no offset} & \textbf{0.643} & \textbf{0.0162} & \textbf{0.0122} \\
No rubber, offset             & $-1.064$  & $0.0390$ & $0.0297$ \\
Fast rubber only, offset      & $-22.423$ & $0.1314$ & $0.1156$ \\
Slow rubber only, offset      & $-3.906$  & $0.0601$ & $0.0584$ \\
Both rubber, offset           & $-34.474$ & $0.1617$ & $0.1565$ \\
\hline
\end{tabular}
\end{table}


\section{Discussion}
We presented SBTenv, an open-source split-belt treadmill environment in MuJoCo, and validated it against experimental data from a passive rimless wheel on a split-belt treadmill \cite{butterfield_split-belt_2022}. SBTenv captured the experimental relationship between belt-speed difference and walking speed without an empirical correction, while running at high throughput.

\subsection{Compliant Contacts Replace the Empirical Correction}
The rimless wheel walks passively on an SBT by harvesting energy from the treadmill: on the fast belt, belt motion raises the wheel and stores gravitational potential energy, which is released as kinetic energy on the slow belt, offsetting collision losses \cite{butterfield_split-belt_2022}. The angular spoke offsets and the BSD together make this exchange possible. Collisions between spokes and the fast belt, however, cause bouncing both experimentally and in simulation. Prior simulations aligned with experiment only after an empirical $4.9$ mm offset of the fast belt in the $z$-direction \cite{butterfield_split-belt_2022}. In SBTenv, compliant spoke contacts, which also more closely reflect the rubber strips on RW-exp, made this correction unnecessary.

The ablation suggests why. Contacts on both spokes without the offset gave the best fit, outperforming the offset-only configuration that most closely reproduces the correction used in \cite{butterfield_split-belt_2022} ($R^2=0.643$ vs. $-1.064$). Combining the two gave the worst fit of all ($R^2=-34.474$). This pattern is consistent with both corrections compensating for the same excess energy retained after spoke-belt collisions: the contacts dissipate this energy through damping, whereas the offset reduces it by altering the contact geometry. Applying both over-corrects, and every such configuration under-predicts the experimental velocity.

\subsection{Measuring Average Steady Velocity}
Butterfield et al. express the RW average steady velocity in terms of spoke contact times \cite{butterfield_split-belt_2022}:
\begin{equation}
{}^{S}v_{\mathrm{avg}}=\frac{2l(\sin{\alpha}+\sin{\beta})-T_{\mathrm{F}}\Delta v_{\mathrm{TM}}}{T_{\mathrm{S}}+T_{\mathrm{F}}}
\label{svavg}
\end{equation}
where ${}^{S}v_{\mathrm{avg}}$ is the \stdv, $l$ is the spoke length, $\alpha$ and $\beta$ are the angular offsets, \dvtm is the BSD, and $T_{\mathrm{S}}$ and $T_{\mathrm{F}}$ are the spoke contact times with the slow and fast belts. In simulation, residual bouncing produces gaps of variable length in the contact signal. Thresholding to fill these gaps is sensitive to outliers, and smoothing can erase true gaps, for example when no spoke touches the belt during rotation. We therefore computed average steady velocity directly from axle displacement over elapsed time. This differs slightly from the experimental procedure, in which RW-exp walked a fixed distance that was divided by its walking time \cite{butterfield_split-belt_2022}, but it avoids estimating contact times.

\subsection{Limitations}
Experimental data were not available in digital form and were digitized from published figures, which introduces measurement error. The rubber contacts and treadmill belts of SBWM-exp were not fully specified, so friction, stiffness, and damping parameters were approximated. The compliant contacts were idealized as spheres ($20.6$ g each) on spring-mass dampers, giving RW-env a total mass of $3.90$ kg compared with $3.53$ kg for RW-exp, although its moment of inertia about the axle remained within $3\%$ of RW-exp ($0.0966$ vs. $0.094$ kg$\cdot$m$^2$). We did not reduce the contact masses to values representative of the thin rubber strips on RW-exp \cite{butterfield_split-belt_2022}, because the spring-mass damper became numerically unstable at the $0.01$ s timestep. Finally, validation used a single passive walker; actively controlled and musculoskeletal walkers will introduce contact and actuation dynamics not tested here.

\section{Conclusion}
We present SBTenv, an open-source, high-throughput, and literature-validated split-belt treadmill environment in MuJoCo. SBTenv reproduces the experimental behavior of a passive rimless wheel on a split-belt treadmill and, through compliant contact modeling, removes the empirical correction required in prior simulation. Because MuJoCo hosts both robot models and converted musculoskeletal models \cite{ikkala_converting_2021}\cite{wang_myosim_2022}, SBTenv opens split-belt walking to the learning-based and optimization-based approaches used to study locomotor adaptation \cite{price_minimum_2023}\cite{seethapathi2024exploration}\cite{fleischmann2025model}.

\section*{AI Usage Disclosure}
The initial development of SBTenv (architecture, validation methods, and ideation) was performed without the use of generative AI. Recent development of SBTenv used Claude Code (Opus 5.5) to assist in refactoring, visualization, feature prototyping, and creation of test cases. All code was reviewed by a human; the human authors are responsible for the content of this manuscript. AI tools were limited in use for the preparation of this manuscript, only for consistency and formatting purposes.

\section*{Acknowledgment}
The authors would like to thank the Sapere Aude Fund, the Massachusetts Institute of Technology Department of Brain and Cognitive Sciences, and the Fairfield University School of Engineering and Computing.

\ifCLASSOPTIONcaptionsoff
  \newpage
\fi
\bibliographystyle{IEEEtran}
\bibliography{references}

\begin{thebibliography}{10}
\providecommand{\url}[1]{#1}
\csname url@samestyle\endcsname
\providecommand{\newblock}{\relax}
\providecommand{\bibinfo}[2]{#2}
\providecommand{\BIBentrySTDinterwordspacing}{\spaceskip=0pt\relax}
\providecommand{\BIBentryALTinterwordstretchfactor}{4}
\providecommand{\BIBentryALTinterwordspacing}{\spaceskip=\fontdimen2\font plus
\BIBentryALTinterwordstretchfactor\fontdimen3\font minus
  \fontdimen4\font\relax}
\providecommand{\BIBforeignlanguage}[2]{{%
\expandafter\ifx\csname l@#1\endcsname\relax
\typeout{** WARNING: IEEEtran.bst: No hyphenation pattern has been}%
\typeout{** loaded for the language `#1'. Using the pattern for}%
\typeout{** the default language instead.}%
\else
\language=\csname l@#1\endcsname
\fi
#2}}
\providecommand{\BIBdecl}{\relax}
\BIBdecl

\bibitem{reisman_neurophysiologic_2010}
\BIBentryALTinterwordspacing
D.~S. Reisman, A.~J. Bastian, and S.~M. Morton, ``Neurophysiologic and
  {Rehabilitation} {Insights} {From} the {Split}-{Belt} and {Other} {Locomotor}
  {Adaptation} {Paradigms},'' \emph{Physical Therapy}, vol.~90, no.~2, pp.
  187--195, Feb. 2010. [Online]. Available:
  \url{https://doi.org/10.2522/ptj.20090073}
\BIBentrySTDinterwordspacing

\bibitem{roemmich_closing_2018}
R.~T. Roemmich and A.~J. Bastian, ``\BIBforeignlanguage{eng}{Closing the
  {Loop}: {From} {Motor} {Neuroscience} to {Neurorehabilitation}},''
  \emph{\BIBforeignlanguage{eng}{Annual Review of Neuroscience}}, vol.~41, pp.
  415--429, Jul. 2018.

\bibitem{price_minimum_2023}
\BIBentryALTinterwordspacing
M.~Price, M.~E. Huber, and W.~Hoogkamer, ``Minimum effort simulations of
  split-belt treadmill walking exploit asymmetry to reduce metabolic energy
  expenditure,'' \emph{Journal of Neurophysiology}, vol. 129, no.~4, pp.
  900--913, Apr. 2023. [Online]. Available:
  \url{https://journals.physiology.org/doi/full/10.1152/jn.00343.2022}
\BIBentrySTDinterwordspacing

\bibitem{seethapathi2024exploration}
N.~Seethapathi, B.~C. Clark, and M.~Srinivasan, ``Exploration-based learning of
  a stabilizing controller predicts locomotor adaptation,'' \emph{Nature
  Communications}, vol.~15, no.~1, p. 9498, 2024.

\bibitem{fleischmann2025model}
S.~Fleischmann, J.~Shanbhag, J.~Miehling, S.~Wartzack, C.~Ong, B.~M. Eskofier,
  and A.~D. Koelewijn, ``A model of the cerebellum generates gait adaptations
  in a reflex-based neuromusculoskeletal model during split-belt walking,''
  \emph{Journal of NeuroEngineering and Rehabilitation}, vol.~22, no.~1, p.
  256, 2025.

\bibitem{todorov2012mujoco}
E.~Todorov, T.~Erez, and Y.~Tassa, ``Mujoco: A physics engine for model-based
  control,'' in \emph{2012 IEEE/RSJ international conference on intelligent
  robots and systems}.\hskip 1em plus 0.5em minus 0.4em\relax IEEE, 2012, pp.
  5026--5033.

\bibitem{butterfield_split-belt_2022}
\BIBentryALTinterwordspacing
J.~K. Butterfield, S.~N. Simha, J.~M. Donelan, and S.~H. Collins,
  ``\BIBforeignlanguage{en}{The split-belt rimless wheel},''
  \emph{\BIBforeignlanguage{en}{The International Journal of Robotics
  Research}}, vol.~41, no. 11-12, pp. 1043--1076, Sep. 2022. [Online].
  Available: \url{https://doi.org/10.1177/02783649221110260}
\BIBentrySTDinterwordspacing

\bibitem{mcgeer_passive_1990}
T.~McGeer, ``Passive dynamic walking,'' \emph{The International Journal of
  Robotics Research}, vol.~9, no.~2, pp. 62--82, 1990.

\bibitem{coleman_motions_1997}
M.~J. Coleman, A.~Chatterjee, and A.~Ruina, ``Motions of a rimless spoked
  wheel: a simple three-dimensional system with impacts,'' \emph{Dynamics and
  Stability of Systems}, vol.~12, no.~3, pp. 139--159, 1997.

\bibitem{anderson_dynamic_2001}
F.~C. Anderson and M.~G. Pandy, ``Dynamic optimization of human walking,''
  \emph{Journal of Biomechanical Engineering}, vol. 123, no.~5, pp. 381--390,
  2001.

\bibitem{srinivasan_computer_2006}
M.~Srinivasan and A.~Ruina, ``Computer optimization of a minimal biped model
  discovers walking and running,'' \emph{Nature}, vol. 439, pp. 72--75, 2006.

\bibitem{geyer_muscle-reflex_2010}
H.~Geyer and H.~Herr, ``A muscle-reflex model that encodes principles of legged
  mechanics produces human walking dynamics and muscle activities,'' \emph{IEEE
  Transactions on Neural Systems and Rehabilitation Engineering}, vol.~18,
  no.~3, pp. 263--273, 2010.

\bibitem{song_deep_2021}
\BIBentryALTinterwordspacing
S.~Song, {\L}.~Kidzi{\'n}ski, X.~B. Peng, C.~Ong, J.~Hicks, S.~Levine, C.~G.
  Atkeson, and S.~L. Delp, ``\BIBforeignlanguage{en}{Deep reinforcement
  learning for modeling human locomotion control in neuromechanical
  simulation},'' \emph{\BIBforeignlanguage{en}{Journal of NeuroEngineering and
  Rehabilitation}}, vol.~18, no.~1, p. 126, Aug. 2021. [Online]. Available:
  \url{https://doi.org/10.1186/s12984-021-00919-y}
\BIBentrySTDinterwordspacing

\bibitem{schumacher_emergence_2025}
P.~Schumacher, T.~Geijtenbeek, V.~Caggiano, V.~Kumar, S.~Schmitt, G.~Martius,
  and D.~F.~B. Haeufle, ``Emergence of natural and robust bipedal walking by
  learning from biologically plausible objectives,'' \emph{iScience}, vol.~28,
  no.~4, p. 112203, 2025.

\bibitem{ong_predicting_2019}
C.~F. Ong, T.~Geijtenbeek, J.~L. Hicks, and S.~L. Delp, ``Predicting gait
  adaptations due to ankle plantarflexor muscle weakness and contracture using
  physics-based musculoskeletal simulations,'' \emph{PLOS Computational
  Biology}, vol.~15, no.~10, p. e1006993, 2019.

\bibitem{nowakowski_deep_2022}
\BIBentryALTinterwordspacing
K.~Nowakowski, K.~El~Kirat, and T.-T. Dao, ``\BIBforeignlanguage{en}{Deep
  reinforcement learning coupled with musculoskeletal modelling for a better
  understanding of elderly falls},'' \emph{\BIBforeignlanguage{en}{Medical \&
  Biological Engineering \& Computing}}, vol.~60, no.~6, pp. 1745--1761, Jun.
  2022. [Online]. Available: \url{https://doi.org/10.1007/s11517-022-02567-3}
\BIBentrySTDinterwordspacing

\bibitem{hwangbo_learning_2019}
J.~Hwangbo, J.~Lee, A.~Dosovitskiy, D.~Bellicoso, V.~Tsounis, V.~Koltun, and
  M.~Hutter, ``Learning agile and dynamic motor skills for legged robots,''
  \emph{Science Robotics}, vol.~4, no.~26, p. eaau5872, 2019.

\bibitem{delp_opensim_2007}
S.~L. Delp, F.~C. Anderson, A.~S. Arnold, P.~Loan, A.~Habib, C.~T. John,
  E.~Guendelman, and D.~G. Thelen, ``{OpenSim}: Open-source software to create
  and analyze dynamic simulations of movement,'' \emph{IEEE Transactions on
  Biomedical Engineering}, vol.~54, no.~11, pp. 1940--1950, 2007.

\bibitem{seth_opensim_2018}
\BIBentryALTinterwordspacing
A.~Seth, J.~L. Hicks, T.~K. Uchida, A.~Habib, C.~L. Dembia, J.~J. Dunne, C.~F.
  Ong, M.~S. DeMers, A.~Rajagopal, M.~Millard, S.~R. Hamner, E.~M. Arnold,
  J.~R. Yong, S.~K. Lakshmikanth, M.~A. Sherman, J.~P. Ku, and S.~L. Delp,
  ``\BIBforeignlanguage{en}{{OpenSim}: {Simulating} musculoskeletal dynamics
  and neuromuscular control to study human and animal movement},''
  \emph{\BIBforeignlanguage{en}{PLOS Computational Biology}}, vol.~14, no.~7,
  p. e1006223, Jul. 2018. [Online]. Available:
  \url{https://journals.plos.org/ploscompbiol/article?id=10.1371/journal.pcbi.1006223}
\BIBentrySTDinterwordspacing

\bibitem{ikkala_converting_2021}
A.~Ikkala and P.~H{\"a}m{\"a}l{\"a}inen, ``Converting biomechanical models from
  {OpenSim} to {MuJoCo},'' in \emph{Converging Clinical and Engineering
  Research on Neurorehabilitation IV}, ser. Biosystems \& Biorobotics.\hskip
  1em plus 0.5em minus 0.4em\relax Springer, 2021, vol.~28, pp. 277--281.

\bibitem{wang_myosim_2022}
H.~Wang, V.~Caggiano, G.~Durandau, M.~Sartori, and V.~Kumar, ``{MyoSim}: Fast
  and physiologically realistic {MuJoCo} models for musculoskeletal and
  exoskeletal studies,'' in \emph{2022 IEEE International Conference on
  Robotics and Automation (ICRA)}, 2022.

\bibitem{tassa_dm_control_2020}
\BIBentryALTinterwordspacing
Y.~Tassa, S.~Tunyasuvunakool, A.~Muldal, Y.~Doron, P.~Trochim, S.~Liu,
  S.~Bohez, J.~Merel, T.~Erez, T.~Lillicrap, and N.~Heess, ``dm\_control:
  {Software} and {Tasks} for {Continuous} {Control},'' \emph{Software Impacts},
  vol.~6, p. 100022, Nov. 2020, arXiv:2006.12983 [cs.RO]. [Online]. Available:
  \url{http://arxiv.org/abs/2006.12983}
\BIBentrySTDinterwordspacing

\bibitem{towers_gymnasium_2025}
\BIBentryALTinterwordspacing
M.~Towers, A.~Kwiatkowski, J.~Terry, J.~U. Balis, G.~D. Cola, T.~Deleu,
  M.~Goulão, A.~Kallinteris, M.~Krimmel, A.~KG, R.~Perez-Vicente, A.~Pierré,
  S.~Schulhoff, J.~J. Tai, H.~Tan, and O.~G. Younis, ``Gymnasium: {A}
  {Standard} {Interface} for {Reinforcement} {Learning} {Environments},'' Nov.
  2025, arXiv:2407.17032 [cs.LG]. [Online]. Available:
  \url{http://arxiv.org/abs/2407.17032}
\BIBentrySTDinterwordspacing

\bibitem{schindelin_fiji_2012}
\BIBentryALTinterwordspacing
J.~Schindelin, I.~Arganda-Carreras, E.~Frise, V.~Kaynig, M.~Longair,
  T.~Pietzsch, S.~Preibisch, C.~Rueden, S.~Saalfeld, B.~Schmid, J.-Y. Tinevez,
  D.~J. White, V.~Hartenstein, K.~Eliceiri, P.~Tomancak, and A.~Cardona,
  ``\BIBforeignlanguage{en}{Fiji: an open-source platform for biological-image
  analysis},'' \emph{\BIBforeignlanguage{en}{Nature Methods}}, vol.~9, no.~7,
  pp. 676--682, Jul. 2012. [Online]. Available:
  \url{https://www.nature.com/articles/nmeth.2019}
\BIBentrySTDinterwordspacing

\bibitem{harris_array_2020}
\BIBentryALTinterwordspacing
C.~R. Harris, K.~J. Millman, S.~J. van~der Walt, R.~Gommers, P.~Virtanen,
  D.~Cournapeau, E.~Wieser, J.~Taylor, S.~Berg, N.~J. Smith, R.~Kern, M.~Picus,
  S.~Hoyer, M.~H. van Kerkwijk, M.~Brett, A.~Haldane, J.~F. del Río, M.~Wiebe,
  P.~Peterson, P.~Gérard-Marchant, K.~Sheppard, T.~Reddy, W.~Weckesser,
  H.~Abbasi, C.~Gohlke, and T.~E. Oliphant, ``\BIBforeignlanguage{en}{Array
  programming with {NumPy}},'' \emph{\BIBforeignlanguage{en}{Nature}}, vol.
  585, no. 7825, pp. 357--362, Sep. 2020. [Online]. Available:
  \url{https://www.nature.com/articles/s41586-020-2649-2}
\BIBentrySTDinterwordspacing

\end{thebibliography}
\end{document}